\documentclass[10pt,twocolumn,letterpaper,pagebackref,breaklinks,colorlinks,allcolors=wacvblue]{article}
\usepackage[pagenumbers]{wacv} 

\usepackage{graphicx}
\usepackage{amssymb}
\usepackage{mathrsfs}

\usepackage{booktabs}
\usepackage{svg}
\usepackage{subfiles}
\usepackage{multirow}
\usepackage{amsmath}
\usepackage{graphicx}
\usepackage{ mathrsfs }
\usepackage{tikz}
\usepackage{todonotes}
\usepackage{rotating}
\usepackage{algorithm}
\usepackage{import}
\usepackage{ amssymb }
\usepackage{graphicx}\usepackage{bbold}
\usepackage{mathtools}
\usepackage{pifont,xcolor}
\usepackage{svg}
\usepackage{orcidlink}
\usetikzlibrary{arrows.meta, decorations.pathmorphing, decorations.markings, shapes.misc}
\definecolor{forestgreen}{RGB}{34,139,34}
\definecolor{olivegreen}{RGB}{107,142,35}
\definecolor{limegreen}{RGB}{50,205,50}
\newcommand{\cmark}{\textcolor{forestgreen}{\ding{51}}}
\newcommand{\xmark}{\textcolor{red}{\ding{55}}}
\usepackage{orcidlink}
\usepackage[noend]{algpseudocode}
\usepackage[T1]{fontenc}
\usepackage{wrapfig}
\usepackage{booktabs}
\usepackage{amsthm}
\usepackage{adjustbox}

\usepackage{amsthm}

\definecolor{wacvblue}{rgb}{0.21,0.49,0.74}
\usepackage[capitalize]{cleveref}

\newcommand{\tfm}{Temporal Flow Matching}
\newcommand{\TFM}{TFM}

\newcommand{\ib}{Last Context Image}
\newcommand{\IB}{LCI}

\newcommand{\dm}{Difference Modeling}
\crefname{section}{Sec.}{Secs.}
\Crefname{section}{Section}{Sections}
\Crefname{table}{Table}{Tables}
\crefname{table}{Tab.}{Tabs.}

\def\wacvPaperID{686} 
\def\confName{WACV}
\def\confYear{2026}

\begin{document}

\title{Temporal Flow Matching for Learning Spatio-Temporal Trajectories in 4D Longitudinal  Medical Imaging}

\author{
Nico Albert Disch ${}^{1,2,3}$\orcidlink{0000-0001-8791-622x}\and
\and Yannick Kirchhoff ${}^{1,2,3}$ \orcidlink{0000-0001-8124-8435}
\and Robin Peretzke ${}^{1,5}$ \orcidlink{0000-0002-6187-3636}
\and Maximilian Rokuss ${}^{1,3}$ \orcidlink{0009-0004-4560-0760}
\and Saikat Roy ${}^{1,3}$ \orcidlink{0000-0002-0809-6524}
\and Constantin Ulrich ${}^{1,5}$ \orcidlink{0000-0003-3002-8170}
\and David Zimmerer ${}^{1,2}$ \orcidlink{0000-0002-8865-2171}
\and Klaus Maier-Hein ${}^{1,2,4,6}$\orcidlink{0000-0002-6626-2463} \\
\\
${}^{1}$ Division of Medical Image Computing, German Cancer Research Center, Heidelberg, Germany  \\
${}^{2}$ HIDSS4Health - Helmholtz Information and Data Science School for Health,\\Karlsruhe/Heidelberg, Germany \\
${}^{3}$ Faculty of Mathematics and Computer Science, University of Heidelberg Heidelberg, Germany \\
${}^{4}$ Pattern Analysis and Learning Group, Department of Radiation Oncology Heidelberg University Hospital\\ Heidelberg, Germany \\
${}^{5}$ Medical Faculty Heidelberg, University of Heidelberg, Heidelberg, Germany \\
${}^{6}$ Pattern Analysis and Learning Group, Department of Radiation Oncology, Heidelberg University Hospital
\\
 \\
{\tt\small nico.disch@dkfz-heidelberg.de}
}

\maketitle

\begin{abstract}
Understanding temporal dynamics in medical imaging is crucial for applications such as disease progression modeling, treatment planning and anatomical development tracking.
However, most deep learning methods either consider only single temporal contexts, or focus on tasks like classification or regression, limiting their ability for fine-grained spatial predictions.
While some approaches have been explored, they are often limited to single timepoints, specific diseases or have other technical restrictions.
To address this fundamental gap, we introduce Temporal Flow Matching (TFM), a unified generative trajectory method that (i) aims to learn the underlying temporal distribution, (ii) by design can fall back to a nearest image predictor, i.e. predicting the last context image (LCI), as a special case, and (iii) supports $3D$ volumes, multiple prior scans, and irregular sampling.
Extensive benchmarks on three public longitudinal datasets show that TFM consistently surpasses spatio-temporal methods from natural imaging, establishing a new state-of-the-art and robust baseline for $4D$ medical image prediction.
 \footnote{Code will be published at \url{https://github.com/MIC-DKFZ/Temporal-Flow-Matching}}
\end{abstract}
\section{Introduction}\label{sec:introduction}
\begin{figure*}[ht]
\centering

\newcommand{\imgsize}{3.5cm}
\rotatebox{90}{\makebox[3.6cm][c]{\large\textbf{ACDC \cite{bernardDeepLearningTechniques2018}}}}
\hspace{0.2cm}
\includegraphics[width=\imgsize,height=\imgsize]{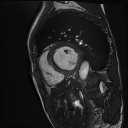}%
\hspace{0.2cm}%
\includegraphics[width=\imgsize,height=\imgsize]{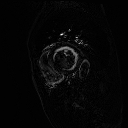}%
\hspace{0.2cm}%
\includegraphics[width=\imgsize,height=\imgsize]{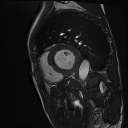}\\[0.5em]

\rotatebox{90}{\makebox[3.6cm][c]{\large\textbf{LUMIERE \cite{suterLUMIEREDatasetLongitudinal2022}}}}%
\hspace{0.24cm}
\includegraphics[width=\imgsize,height=\imgsize]{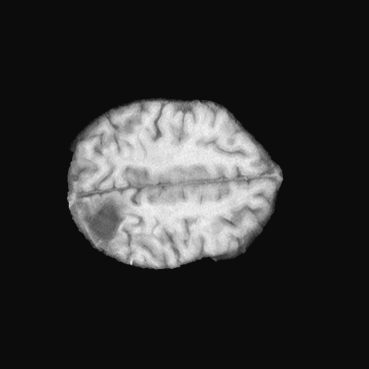}%
\hspace{0.2cm}%
\includegraphics[width=\imgsize,height=\imgsize]{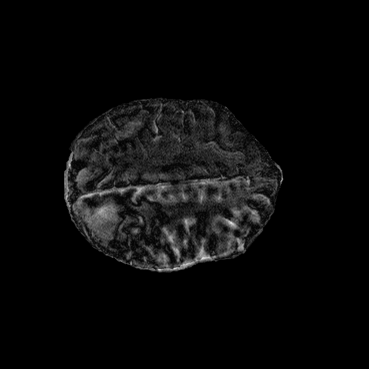}%
\hspace{0.2cm}%
\includegraphics[width=\imgsize,height=\imgsize]{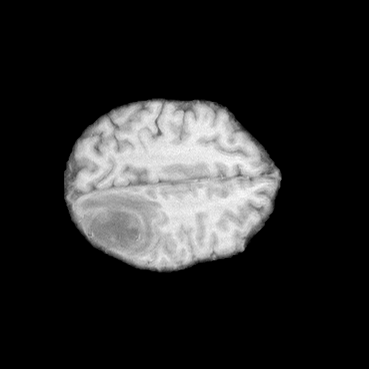}\\[0.5em]

\rotatebox{90}{\makebox[3.6cm][c]{\large\textbf{ISLES \cite{riedelISLES2024First2024}}}}%
\hspace{0.2cm}
\includegraphics[width=\imgsize,height=\imgsize]{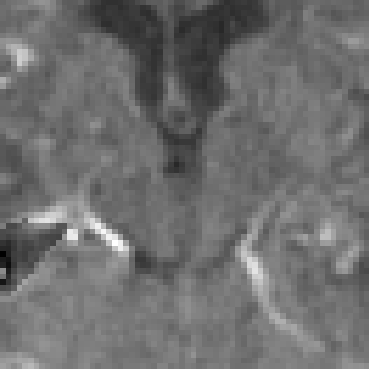}%
\hspace{0.2cm}%
\includegraphics[width=\imgsize,height=\imgsize]{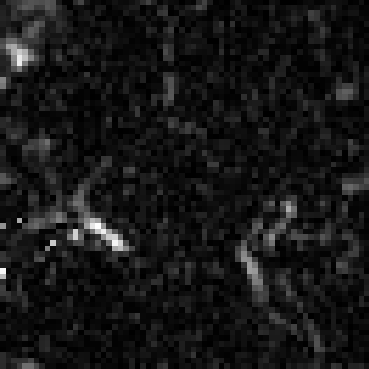}%
\hspace{0.2cm}%
\includegraphics[width=\imgsize,height=\imgsize]{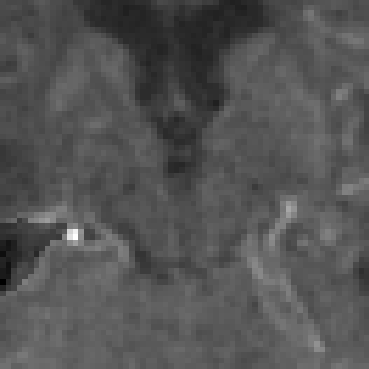}\\[0.5em]

\hspace*{0.5cm}%
\makebox[\imgsize]{\large Target (GT)}%
\hspace{0.1cm}%
\makebox[\imgsize]{\large Difference to \IB{}}%
\hspace{0.1cm}%
\makebox[\imgsize]{\large  Last Context \IB{}}

\caption{\textbf{The last available context image is close to the target image}. Rows show different subjects; columns show target, difference, and \ib{}(\IB{}).
The upper row shows the ACDC dataset, which is a dataset about cardiac imaging, with different pathologies.
These different pathologies manifest in different ways, such as different heart cycles.
The middle row shows the Lumiere dataset, which is a dataset about brain tumors with follow up scans after treatment.
Scans occur after a few weeks, and the brains are extracted from the skull and registered.
The last row shows the ISLES dataset, which is a dataset about stroke imaging.
The images show the perfusion CT, which shows the vessels in the brain.
For better visualization, a center crop is shown. The motivation is that \IB{} is very close to the target, and using that as a prior yields quick results. 
}
\label{fig:dataset-grid}
\end{figure*}
Longitudinal medical imaging is essential for tracking disease progression, monitoring treatment effects, and modeling anatomical development.
When a patient undergoes imaging across multiple visits, whether for disease monitoring or post-treatment follow-ups, a longitudinal series is created.
Moreover, there are multiple modalities which intrinsically contain temporal dimensions, such as ultrasound, Cine-MRI or perfusion CT.
Despite the inherent temporal structure of such data, most current deep learning approaches analyze images as isolated time points, ignoring the valuable temporal dimension.
Applications of longitudinal imaging span a wide range of clinical tasks, including neurodegenerative disease progression (e.g.\ Alzheimer’s disease~\cite{petersenAlzheimersDiseaseNeuroimaging2010}), cardiac motion analysis~\cite{bernardDeepLearningTechniques2018}, and treatment response prediction in oncology~\cite{suterLUMIEREDatasetLongitudinal2022,calabreseUniversityCaliforniaSan2022}.
However, deep learning for spatio-temporal medical imaging remains underexplored compared to image analysis approaches that focus on single timepoints.
Most existing approaches focus on classification and regression such as~\cite{zeghlacheLaTiMLongitudinalRepresentation2024,zhangM2FusionMultitimeMultimodal2024}.
Albeit valuable, these tasks do not fully represent fine-grained changes in the images.
High-dimensional generative models offer richer insights, as they can model the evolution of structures like tumors over time rather than merely detecting changes.
Generative models, such as diffusion models~\cite{litricoTADMTemporallyAwareDiffusion2024,yuanLongitudinallyConsistentIndividualized2024,puglisiEnhancingSpatiotemporalDisease2024} and Neural ODEs~\cite{lachinovLearningSpatioTemporalModel2022,liuImageFlowNetForecastingMultiscale2025a}, have been applied to medical imaging, but they also predominantly operate on single time points, having only partially available context, restricting their applicability.
Some approaches embed multiple time points~\cite{baiNODERImageSequence2024}, yet they still only encode single images independently. 
In contrast, jointly using multiple observations has been shown to enhance prediction accuracy~\cite{fangDeepLearningPredicting2021}.
Other approaches interpolate images between two time points~\cite{zhuLoCIDiffComLongitudinalConsistencyInformed2024}, limiting their use for predictive purposes.
Consequently, current techniques are either technically constrained, limiting their general application to longitudinal imaging, or rely on disease-specific priors.
\begin{table*}[!h]
\centering
\begin{tabular}{llcccc}
\toprule
Category & Method & $3D$ & Disease Agnostic & Multiple Contexts & \dm* \\
\midrule
\multirow{3}{*}{Medical Imaging} 
  & NODER~\cite{baiNODERImageSequence2024} & \cmark & \cmark & \xmark & \xmark \\
  & Image Flow~\cite{liuImageFlowNetForecastingMultiscale2025a} & \xmark & \cmark & \xmark & \cmark \\
  & BrLP~\cite{puglisiBrainLatentProgression2025} & \cmark & \xmark & \xmark & \xmark \\
\midrule
\multirow{3}{*}{Natural Imaging}
  & ConvLSTM~\cite{shiConvolutionalLSTMNetwork2015} & \cmark & \cmark & \cmark & \xmark \\
  & SimVP~\cite{gaoSimVPSimplerBetter2022} & \cmark & \cmark & \cmark & \xmark \\
  & ViViT~\cite{arnabViViTVideoVision2021} & \cmark & \cmark & \cmark & \xmark \\
\midrule
\multirow{1}{*}{Ours}
  & \TFM{} & \cmark & \cmark & \cmark & \cmark \\
\bottomrule
\end{tabular}
\caption{\textbf{Technical comparison of spatio-temporal prediction methods.} Methods are grouped by origin (medical or natural imaging). \TFM{} satisfies all requirements. *\dm indicates modeling changes from context instead of full images.}
\label{tab:comparison}
\end{table*}

 \\ \\ \noindent
Yet our experiments demonstrate that  spatio-temporal methods from natural imaging cannot outperform a simple baseline: \ib{} (\IB{}), which used the most recent image as a prediction.
Table \ref{tab:comparison} summarizes technical comparison of baselines from medical and natural imaging:
\textbf{Static Bias:} Pixel level scores are dominated by unchanged anatomy. Figure \ref{fig:dataset-grid} shows the differences between consecutive frames. We note that changes are small, and in some cases quite localized. Full dataset statistics can be found in Figure \ref{fig:ratios_mse_lci}. For example, in the ACDC dataset~\cite{bernardDeepLearningTechniques2018} see e.g. Figure~\ref{fig:dataset-grid}, the temporal differences account for only $\sim 3\%$ of $NRMSE$.
\\ \\ \noindent
Motivated by these observations, we introduce  \textbf{\tfm{}} (\textbf{\TFM{}}), a unified generative trajectory model that captures $3D$ temporal evolution across multiple scans, modeling only the changes.
We term this mechanism as \textbf{\dm.}
Crucially, this modeling objective imposes no architectural or regularization constraints, since it is mathematically just a transformation of the output space (see Appendix \ref{sec:appendix-a_semantics} for further discussion). 
Therefore, \TFM{} remains fully flexible and offers the following capabilities:
\begin{itemize}
    \item \textbf{Efficient Training}: Offers end-to-end optimization within $11.3GB$ during training
    \item \textbf{4D Time Series} Handles 3D volumetric time series of variable length and amount of context
    \item \textbf{Robust to Sparse and Irregular Sampling} Robust to irregular or missing follow-up scans
    \item \textbf{Disease and Modality Agnostic} Generalizes across heterogeneous applications, including cardiac function (Cine-MRI), stroke progression  (perfusion CT) and glioblastoma growth (MRI).
\end{itemize}
\noindent
Through extensive benchmarks on three public longitudinal and spatio-temporal datasets, \TFM{} consistently outperforms our prior spatio-temporal baseline, including \IB{}.
To the best of our knowledge, this results in the first comprehensive benchmark of spatio-temporal prediction methods in medical imaging.
With its  strong performance and broad technical flexibility, \TFM{} establishes a robust foundation and  new baseline for future advances in $4D$  medical image analysis.

\section{Methods}\label{sec:methods}

Longitudinal medical imaging requires handling of irregularly sampled time series, while capturing spatial and temporal dynamics.
In Section~\ref{subsec:problem_setup}, we formalize the problem of irregular medical imaging.
We  summarize Flow Matching (FM) in Section~\ref{subsec:fm}, and discuss challenges with integrating FM into image time series.
We then introduce a novel extension of FM, namely \TFM{}, in Section~\ref{subsec:tfm}, designed to explicitly address these challenges.
Finally, in Section~\ref{sec:sparsity_filling}, we address missing images by introducing a  sparsity filling strategy, which is essential for maximizing the performance of \TFM{}.
\begin{figure*}[htb]
    \centering
    \includesvg[width=0.99\textwidth]{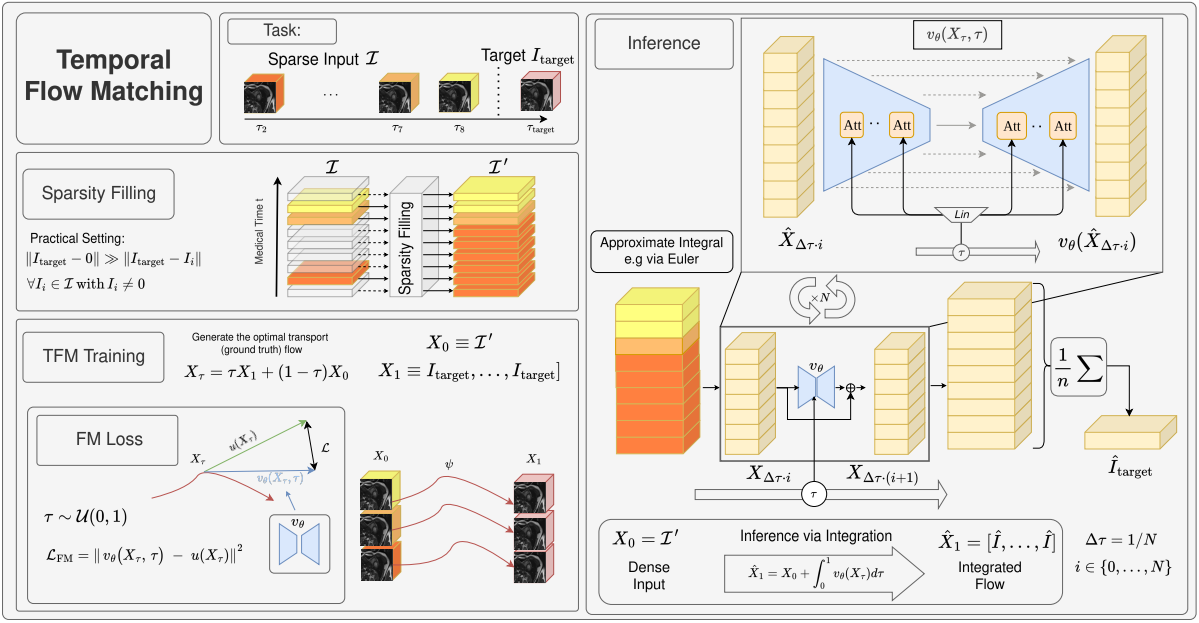} 
    \caption{\textbf{Method and Modeling overview of \tfm{}}.
The method starts with \textbf{Sparsity Filling}, where we fill sparse image sequences $\mathcal{I}$ to a filled input $\mathcal{I}'$.
This prevents huge variations in flow size, with a similar motivation as to why we use FM in the first place.
During \textbf{Training}, we apply the standard FM objective. 
We generate a ground truth velocity field via linear interpolation and train the network $v_\theta$ to predict this velocity at each FM step $\tau$ for the intermediate state $X_\tau$.
The inset shows the architecture of $v_\theta$.
In \textbf{Inference}, we approximate the continuous flow by integrating $v_\theta$ (illustrated here with Euler), to transport context frames towrd the target image.
}
    \label{fig:Figure1}
\end{figure*}
\subsection{Problem Setup}
\label{subsec:problem_setup}
Let us assume a dataset of $p$  spatio-temporal image sequences  (i.e.\, one per patient).
For each patient,  we assume  $T$ \textbf{context} images $\mathcal{I} = \{ I_1, \dots, I_T \}$ with $I_i \in \mathbb{R}^{H\times D \times W}$ acquired at ordered, and possibly irregular,time points $\mathcal{T} = \{ t_1, \dots, t_T \}$,  with a \textbf{target} image $I_\text{target}$ at a time $t_\text{target}$.
Due to irregular and sparse acquisitions, missing context images are set to $0$.
For this task, we propose \textbf{Temporal Flow Matching (TFM)}, a generative model that extends \textbf{Flow Matching (FM)} to predict future medical images from sparse and irregular historical observations. 
\subsection{Flow Matching (FM)}
\label{subsec:fm}
We adopt the notation of Flow Matching (FM)  as introduced in~\cite{lipmanFlowMatchingGenerative2023}.
FM learns a continuous transformation between a source sample $X_0\sim p$ and a target sample  $X_1 \sim q$ by modeling an \textbf{optimal transport field} $\psi$ parametrized by an Ordinary Differential Equation (ODE):
\begin{equation}
\label{eq:integration}
    \frac{d}{d\tau} \psi_\tau(x) = u_\tau(\psi_\tau(x)), \quad \text{with  }\,  X_1 = X_0 + \int_0^1 u_\tau(X_\tau) d\tau,
\end{equation}
where $\psi_\tau(X_0)=X_\tau$ defines the trajectory at interpolation step $\tau \in [0,1]$.
Since the equation~\eqref{eq:integration} is an ODE, we can fix the initial conditions as $X_0$.
The vector field $u_\tau$ denotes the velocity of the transport field $\psi$ at position $\psi_\tau(X_0)$.
To avoid ambiguity with real-valued medical timepoints, we refer to $\tau$ as the \emph{FM step}, rather than the time $t$.
A neural network $v_\theta$ is trained to predict the true velocity field:
\begin{equation}
\label{eq:vel_network_def} 
    v_\theta(X_\tau,\tau) \approx u_\tau,
\end{equation}
where $X_\tau$ is the intermediate state at step $\tau$, and $\theta$ are the network parameters.
As only $X_0$ and $X_1$ are observed, we define $X_\tau$ using a known transport map $\psi$.
Typically, a linear interpolation is used:
\begin{equation}
    X_\tau = (1-\tau)X_0 +\tau X_1,\label{eq:linear_interpolation_fm}
\end{equation}
while other choices are possible.
The FM training objective then minimizes the discrepancy between the predicted velocity and true velocity:
\begin{equation}
\label{eq:fm_training_loss}
    \mathcal{L}_{FM} = \mathbb{E}_{\tau  \sim \mathcal{U}(0,1), X_0\sim p} \|v_\theta(X_\tau,\tau)-u_\tau\|.
\end{equation}
Unlike diffusion models, which rely on iterative denoising guided by learned score functions, FM learns a direct mapping via velocity fields.
Under certain conditions, FM can be shown to be equivalent to diffusion models~\cite{lipmanFlowMatchingGuide2024}.
\subsection{Temporal Flow Matching (TFM)}
\label{subsec:tfm}
\begin{algorithm*}
\caption{Temporal Flow Matching: Training and Inference}
\begin{algorithmic}[1]
  \Require Patients $\mathcal{P} = \{[\mathcal{I}_1 , I_{\text{target},1}], \dots, [\mathcal{I}_p , I_{\text{target},p}]\}$ and initial network $v_\theta$
  \While{training}
    \State $[\mathcal{I} , I_{\text{target}}] \sim \mathcal{P}(\mathcal{X})$ \Comment{pick a random patient}
    \State  $\tau \sim \mathcal{U}(0,1)$ \Comment{pick a random flow step}
    \State $\mathcal{I}_\text{target} \gets [I_\text{target}, \dots, I_\text{target}]$ \Comment{Extend the dimension of $I_\text{target}$ $T$ times}
    \State $\mathcal{I}' \gets \text{Sparsity Filling}(\mathcal{I})$ \Comment{Fill empty images}
    \State $X_\tau \gets (1 - \tau)\,\mathcal{I}' + \tau\,\mathcal{I}_\text{target} $ \Comment{Calculate the linear interpolation between each context and target}
    \State
      $\mathcal{L}_{\mathrm{TFM}} \gets
         \left\|\,v_\theta\bigl(\tau,\,X_\tau\bigr)\;
           -\;(\mathcal{I}_\text{target}-\mathcal{I}')
         \right\|^2$ \Comment{Calculate the velocity loss}
    \State Update $\theta \;\leftarrow\;\mathrm{AdamW} \left(  \nabla_\theta \mathcal{L}_{\mathrm{TFM}} \right)$
  \EndWhile
  \State \Return $v_\theta$
    \If{inference}
\State Initialize $X_0 \gets \mathcal{I}'$
\State Define integration grid $\{\tau_0 = 0, \dots, \tau_N = 1\}$ with $n$ steps 
\State $\hat{X}_{0:N} \gets \mathrm{ODEInt}\big(v_\theta, X_0,\; \{\tau_0, \dots, \tau_N\}\big)$ \Comment{numerically integrate the network}
  \State \Return $\hat{X}_N$
\EndIf
\end{algorithmic}
    \label{alg:tfm}
\end{algorithm*}

Medical image follow-ups are often irregular, both in terms of temporal spacing and the number of available context images.
This poses a challenge for standard generative models, such as Flow Matching (FM) or Diffusion, which models a transformation between two distributions $q$ and $p$.
Therefore, FM cannot be directly applied when the input and target sequences differ in dimensionality.
There are two canonical strategies to address this; 
i): \textbf{Temporal Pooling}: Compress  $\mathcal{I}$ via a spatio-temporal encoder, or predict the flow only from the last available image.
ii): \textbf{Dimension padding} Extend the target and the context dimensionality of the to a set context sequence length. 
We adopt Dimension Padding, since we find that only using flows from the last image is not stable.
The second method is in part inspired by~\cite{gaoSimVPSimplerBetter2022}, which also lifts all predictions to the same temporal dimension as a fixed input dimension.
With this, we propose \textbf{\tfm(\TFM{})}, a generative model that directly learns transformations from \textbf{each} context image to the target image within a unified spatio-temporal flow formulation.
Unlike approaches that compress temporal information early, or operate on latent representation, \TFM{} retains \textbf{full spatial resolution}.
This is feasible because \TFM{} has a computational footprint comparable to other spatio-temporal methods, which makes it able to afford this modeling compute.
By jointly processing the entire input sequence, the model can leverage spatio-temporal dependencies between input images.
To enable this, we define the FM target as
\begin{equation}
    X_1 \coloneqq [\underbrace{I_\text{target}, \dots , I_\text{target}}_T],\label{eq:target_distribution_definition}
\end{equation}
where $X_1 \in \mathbb{R}^{T\times D\times H \times W}$, with $T$ being the number of context images.
The FM initial conditions for equation~\eqref{eq:integration} then reads:
\begin{equation}
\label{eq:tfm}
     X_0 = \mathcal{I} \quad \text{and} \quad X_1 = [\underbrace{I_\text{target}, \dots , I_\text{target}}_T] 
\end{equation}
where $\mathcal{I}$ is the series of input images.
Then we have the vector field $\psi_\tau: \mathbb{R}^{T\times D\times H \times W} \to \mathbb{R}^{T\times D\times H \times W}$.
Training and inference are described in Algorithm~\ref{alg:tfm}.
We then calculate $X_\tau = \psi_\tau(X_0)$ and $u_\tau(x) = \frac{d}{dt} \psi_\tau(x)$ using \eqref{eq:linear_interpolation_fm}.
The \textit{neural net} $v_\theta$ then predicts a velocity $\hat{u}_\tau$~\eqref{eq:vel_network_def}, and is trained via~\eqref{eq:fm_training_loss}.
Inference is then done using eq. \eqref{eq:integration}, i.e. 
\begin{equation}
\label{eq:tfm_final}
\hat{X}_1= X_0 + \int_0^1 v_\theta(X_\tau, \tau) d\tau.
\end{equation}
In practice, equation~\eqref{eq:tfm_final} can only be solved numerically.
This requires choosing an ODE solver (e.g.\, Euler or Runge-Kutta) and the number of integration steps (and optionally solver hyperparameters).
Since $\hat{X}_1 \in  \mathbb{R}^{T\times D\times H \times W}$, we need to reduce the temporal dimension.
For the final temporal reduction, we use either the last predicted time channel or the mean across time.
\paragraph{\dm{}}
Rather than modeling the whole spatio-temporal image distribution, our method predicts the velocity field, meaning the differences between context and target.
Standard Flow Matching transforms between two distributions $p$ and $q$, but here both stem from the same patient at different timepoints.
Consequently, the velocity is the \textit{difference} between $\mathcal{I}'$ and $\mathcal{I}_\text{target}$.
Hence, we call this mechanism \textit{\dm}, since $v_\tau$ \textit{models} this \textit{difference}.
See Appendix ~\ref{sec:appendix_paradox} for a toy example illustrating how this modeling can influence evaluation metrics.
\subsection{Handling Missing Data: Sparsity Filling}\label{subsec:handling-missing-data:-sparsity-filling}
\label{sec:sparsity_filling}
Irregular sampling in longitudinal data creates 'holes' in the time axis (i.e., missing images for certain time points), which can distort the estimated flow between $I_i$ and $I_\text{target}$.
This reflects the same issue discussed in the motivation, but now arising from missing context frames.
To address missing context images, we apply \textbf{sparsity filling}, replacing them with the most recent available scan (see Fig.~\ref{fig:Figure1} for visualization).
This ensures smoother inputs and more stable flow estimation across masked inputs.
If missing frames occur before the first available scan, we fill them using the earliest available image.
We denote the filled context sequence via $\mathcal{I}'$. 
We hypothesize this helps because each filled image in $\mathcal{I}'$ is closer to $I_\text{target}$ than an empty/ zero-filled image, resulting in more homogeneous flow fields.
In out ablation studies, sparsity filling was essential; omitting it leads to unstable training and degraded convergence.
\section{Data and Experimental Design}\label{sec:data-and-experimental-design}
\begin{table*}
\centering
\caption{\textbf{Quantitative Evaluation on Test Sequences:} Reported values are mean (standard deviation) over three runs. 
Metrics include normalized root $MSE$, $NRMSE$, structural similarity index ($SSIM [\%]$) and peak signal-to-noise-ratio $PSNR$.
*ViViT on Lumiere ran out of $40GB$ memory, despite having a smaller batch size and the lowest possible feature size. }
\label{tab:main_results}
\begin{tabular}{lllll}
\toprule
Dataset & Model & NRMSE & SSIM[$\%$] & PSNR  \\
\midrule
\multirow[t]{5}{*}{ACDC}& \IB{} &0.056 & 93.3 & 28.49 \\
& ConvLSTM & 0.112 (0.005) & 50.4 (1.5) & 19.12 (0.31) \\
 & SimVP & 0.124 (0.001) & 52.8 (1.6) & 21.21 (0.13) \\
 & ViViT & 0.120 (0.008) & 30.1 (6.9) & 18.47 (0.53) \\
 & TFM (ours) & \textbf{0.040 (0.012)} & \textbf{94.5 (0.8)} & \textbf{30.51 (1.56)} \\
\cline{1-5}
\multirow[t]{5}{*}{ISLES}& \IB{} & 0.057 & 95.6 & 28.39 \\
& ConvLSTM & 0.182 (0.005) & 40.8 (0.9) & 17.85 (0.23) \\ 
 & SimVP & 0.124 (0.001) & 52.8 (1.6) & 21.21 (0.13) \\
 & ViViT & 0.162 (0.003) & 32.5 (0.8) & 18.84 (0.21) \\
 & TFM (ours) & \textbf{0.041 (0.007)} & \textbf{97.6 (0.8)} & \textbf{31.03 (1.08)} \\
\cline{1-5}
\multirow[t]{4}{*}{Lumiere*} & \IB{}  & 0.085 & 89.3 & 21.55 \\
& ConvLSTM & 0.352 (0.009) & 7.9 (4.2) & 9.12 (0.22) \\ 
 & SimVP & 0.711 (0.028) & -2.5 (0.8) & 2.98 (0.34) \\
 & TFM (ours) & \textbf{0.069 (0.007)} & \textbf{89.7 (1.2)} & \textbf{23.73 (0.82)} \\
\cline{1-5}
\bottomrule
\end{tabular}
\end{table*}
We compare \TFM{}  to methods that jointly model spatial and temporal information across multiple time points.
\textbf{SimVP}~\cite{gaoSimVPSimplerBetter2022}: This method originates from the natural image domain and simply uses all context images as input of their network.
The original architecture consists of a 2D UNet, which we  extend here to 3D\@.
The temporal information is handled via flattening the time dimensions into the channel dimension.
\textbf{ConvLSTM}~\cite{shiConvolutionalLSTMNetwork2015}: It extracts spatial features using convolutions while capturing temporal dependencies through an LSTM’s recurrent states~\cite{hochreiterLongShortTermMemory1997}.
At each time step, the model processes an input image using convolutional layers and updates its internal memory, which maintains information across the sequence.
\textbf{ViViT} The Video Vision Transformer (ViViT) first processes all input context images into image patches, where the patch size is $8\times32\times32$.
We use the ViViT as in ~\cite{yoonSADMSequenceAwareDiffusion2023}, for fair comparison of the pure spatio-temporal backbone.
\paragraph{Baseline Training}
The baseline methods directly predict the target given the context sequence $\mathcal{I}$.
So the loss for those methods reads:
\begin{equation}
    \mathcal{L}_{\text{baseline}} = \|f_\theta(\mathcal{I}) - I_\text{target}  \|.\label{eq:baseline_loss_function}
\end{equation}
Further definitions of \textit{architecture, model and method} is found in section~\ref{sec:appendix-a_semantics}.
\noindent
\paragraph{\ib{}}
Furthermore, we use the \ib{} (\IB{}), a heuristic that serves as an estimated lower bound.
\IB{}  is denoted as the last image in the sequence which is non-zero.
This baseline is medically motivated, as it serves as a part of medical decision making when looking at longitudinal series (see~\cite{therasseNewGuidelinesEvaluate2000}).
\footnote{While \IB{} is optimal for \textit{monotonic} sequences, it might not be the best performing image from the context sequence.
However, selecting the best image from the sequence would require further insight or an oracle model.
So \IB{} is the best we can naively do for most tasks. Yet for the datasets we consider \IB{} is in fact the best.}
\noindent
\subsection{Datasets}\label{subsec:datasets}
\textbf{ACDC}~\cite{bernardDeepLearningTechniques2018} is a cardiac MRI dataset for different states of the heart.
Images are reshaped to $[T,H,D,W]=[11,32,128,128]$, where the target is a single image having the same spatial dimensions.
For the ACDC dataset, we randomly mask out time points, in order to make it irregular.
We split the dataset into $80$ training, $20$ validation and $50$ test images.
This dataset was used for method development and ablations were done on the validation set.
\\ \noindent
\textbf{ISLES} ~\cite{riedelISLES2024First2024} consists of perfusion CT images from stroke patients.
For our experiments, we utilize this 4D modality.
Since there are dozens of time steps with minor changes in the image, we further process the image.
For that, we only take every other time step of the perfusion sequence.
From the resulting series, we randomly pick 4 consecutive points, where the last point is the target, and we randomly mask context images.
The context then has shape $[T,H,D,W]=[7,16,128,128]$.
This dataset is split into $92$ training, $23$ validation and $34$ test images.\\ \noindent
\textbf{Lumiere}~\cite{suterLUMIEREDatasetLongitudinal2022} is a longitudinal dataset of tumor growth in gliomas, consisting of 3D MRI scans.
The images are reshaped to $[T,H,D,W]=[7,96,96,64]$.
Since not all patients have many acquisitions, we \textit{prepended} zeros to ensure pre-processing is consistent.
For Lumiere we have $48$ training, $12$ validation and $14$ test images.
Example images from two timepoints for each dataset are shown in Figure~\ref{fig:dataset-grid}.
\subsection{Experimental settings}\label{subsec:experimental-settings}
All methods (see \ref{sec:appendix-a_semantics} for notation) were trained with AdamW and a cosine-annealed learning-rate schedule, using a batch size of 4.
The learning rate was fixes at $1\mathrm{e}{-4}$ for all experiments.
For \TFM{}, we used 10 integration steps during inference (see Table~\ref{tab:ssim_vs_memory}).
Our \TFM{} builds on the standard UNet from the TorchCFM library~\cite{tongTorchCFM2025}, using cross-attention between time embeddings and spatial feature maps (see Figure~\ref{fig:Figure1} for an overview).
To ensure fair comparison, we ran each experiment three times with different validation splits and the same random seed within each split.
\paragraph{Random Masking}
For ACDC and ISLES, we randomly omit context images during both training and validation, to simulate irregular sampling. 
Since we believe we are the first to benchmark methods in this very specific irregular setting, we \textbf{highlight a potentially grave pitfall}:
If validation masks are resampled at each validation epoch, even with a fixed seed, the masking evaluation metrics change every time, which is exacerbated by our small validation set .
This variability affects even the trivial \IB{} baseline and makes "best" epoch selection arbitrary.
Since the validation set is small, context sequences can be extremely sparse or dense, causing the \IB{} baseline's performance to fluctuate drastically\footnote{In natural imaging, validation sets are much larger, so random fluctuations are less severe. In medical imaging, however, smaller validation sets make these fluctuations significant.}.
To avoid this issue, we generate one fixed set of masks per split (using a single seed) and reuse those exact masks for every model at every validation epoch.
This ensures consistent validation conditions, meaningful epoch selection, and fully reproducible and interpretable validation results.
In all cases, models were selected by the lowest validation $MSE$ and then evaluated on the held-out test set.

\graphicspath{{\subfix{../Figures/}}}
\section{Results and Discussion}\label{sec:results}
\begin{figure}[htbp]
    \centering
    \includegraphics[width=0.95\linewidth]{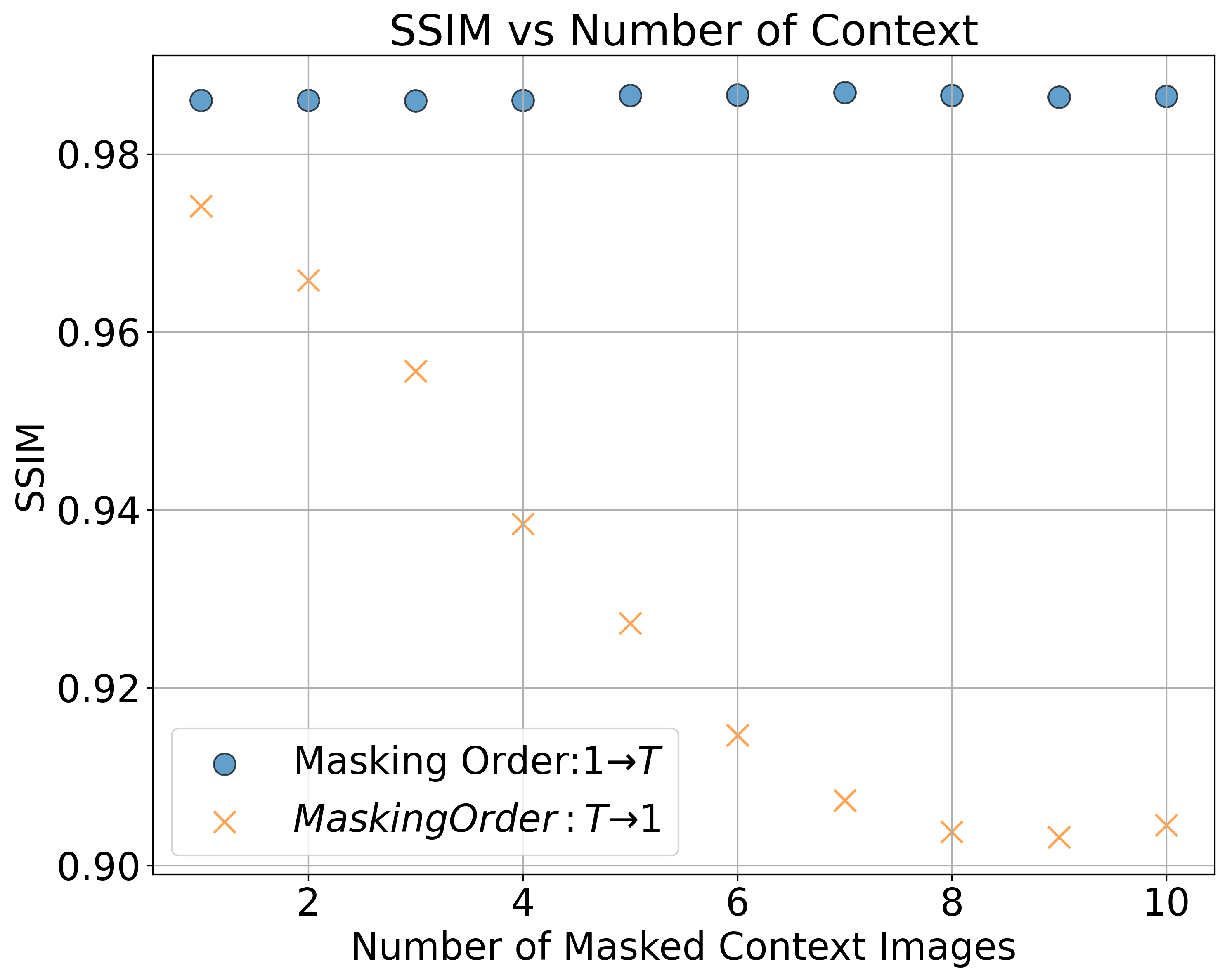}
    \caption{\textbf{Zero-Shot masking of early and late context images:} This plot shows \TFM{}'s prediction quality on ACDC under two masking protocols at inference time. The model was trained under the irregular masked setting on ACDC.
   $1\to T$:  we progressively remove earlier context frames, i.e. the ones which are furthest away from the target.
   $T\to 1$: we remove frames closest to the target. The x-axis indicates the total number of masked frames. 
   As expected, masking later points leads to a steady degradation. 
   For masking earlier points leads to a very slight u curve at the 3rd significant digit, incidentally around $5$ masked points.
   }
    \label{fig:front_back_masking}
\end{figure}
\begin{figure}
    \centering
    \includegraphics[width=0.95\linewidth]{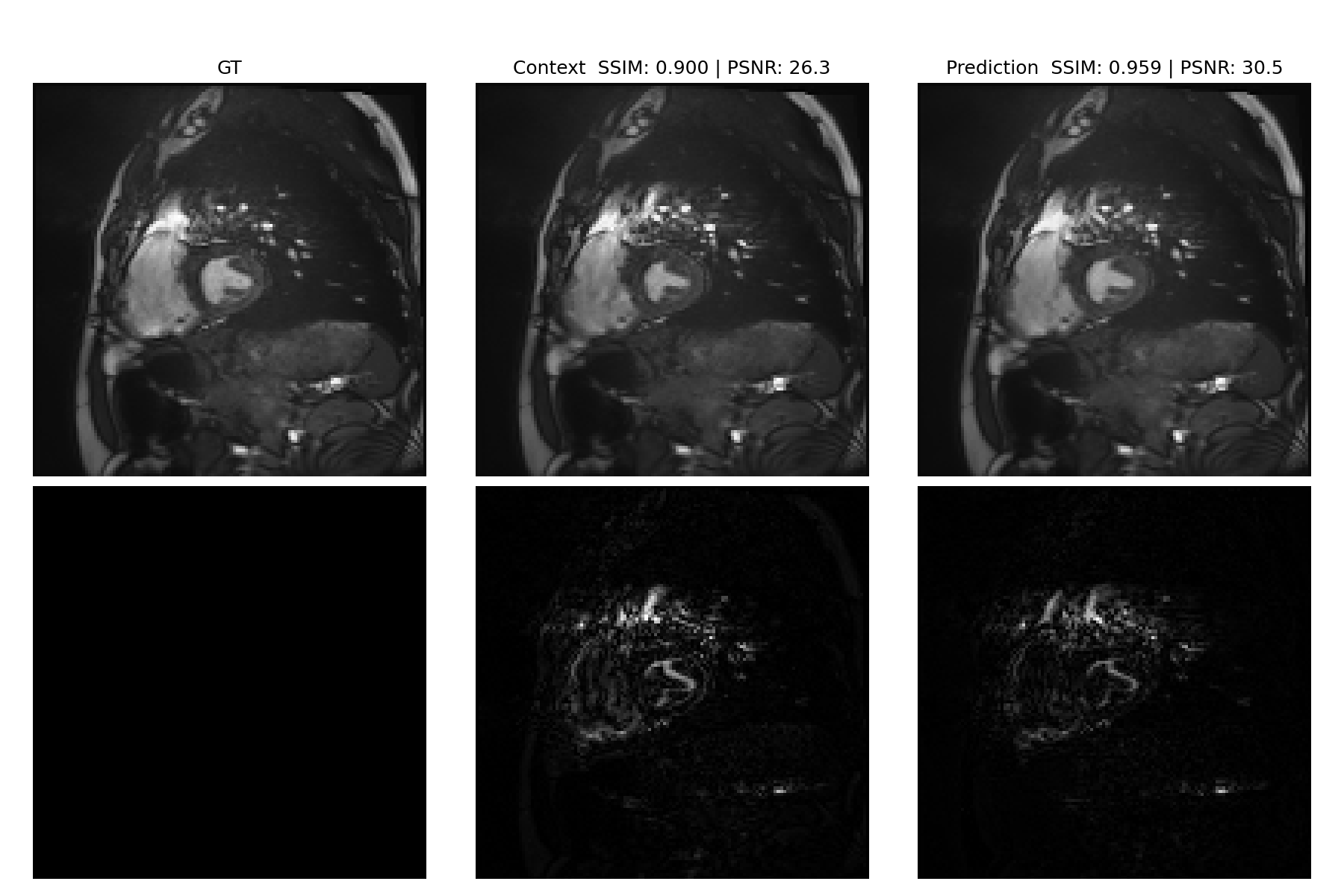}
    \caption{\textbf{Qualitative results} Left: target image. Middle: \IB{}. Right: \TFM prediction. Note how \TFM{} yields a very strong prediction where context and target do not differ, and \TFM{} yields a better prediction where \IB{} differs more strongly from the target. The resulting image is a realistic looking reconstruction.}
    \label{fig:qualitative_results_example}
\end{figure}

\noindent
\textbf{\TFM{} outperforms \IB{}} across all datasets and metrics as shown in Table~\ref{tab:main_results}.
It achieves top performance on every metric and dataset tested.
Competing methods struggle to generate realistic images, often scoring below the \IB{} baseline.
We stress that pixel-wise metrics such as $MSE$ uniformly penalize any change, so unchanged anatomy dominates the score and can mask fine spatio-temporal predictions (see e.g. Figure~\ref{fig:ratios_mse_lci}).
By modeling the difference directly, we argue that \TFM{} sidesteps this bias and more faithfully captures true temporal evolution.
Future work should consider modeling  differences directly, capturing metrics only on regions of interests with substantial change, or adopting task-specific metrics that align more with clinical motivations~\cite{maier-heinMetricsReloadedRecommendations2024}.
\textbf{Lumiere} is a longitudinal tumor growth dataset, characterized by sparse sequences and a small number of training cases.
The strong differences in patient-specific trajectories and its data scarcity make Lumiere particularly difficult; most methods fail to come close to \IB{}, except for \TFM{}.
We attribute the poorer performance of other methods to the small training set size and high inter-subject variability.
This leads to negative $SSIM$ for the SimVP method, and qualitatively noisy results.
Despite these challenges, \TFM{} outperforms \IB{} on Lumiere in both $MSE$ and $PSNR$.
This supports our hypothesis that \TFM{} benefits from \textit{\dm}, making it more robust in real-world scenarios.
These findings suggest that \tfm{} is especially well-suited for real-world scenarios.
Figure~\ref{fig:qualitative_results_example} illustrates that \TFM{} generates realistic images.
Further qualitative results can be found in the appendix.
Thanks to our use of Runge-Kutta integration, memory savings are non-trivial;
memory usage can be further reduced by switching to Euler integration and detaching tensors after every step, or to aim for single step predictions.
This could significantly reduce memory usage, if needed.
\begin{table}[h]
    \centering
        \caption{\textbf{Ablation Results for \TFM{} on ACDC:} 
        This table compares \TFM{} under different design changes, showing the performance under each scenario.
        The ablations were done on an ACDC validation set.
        We evaluate the effect of using a more lightweight version of the UNet which does not use attention('No Att').
        Instead, $\tau$ and image embeddings are merged via concatenation in the bottleneck. 
        We also compare aggregating via the mean and the last image, but these results are only for inference. Training is still done the same way.
        Third, we compare sparsity filling with the alternative of using the image sequences $\mathcal{I}$ as they are given. 
        This notable reduces performance.
         *Limiting the model to only see \IB{} during training and perform FM on this is unstable, which highlights the importance of temporal context.
        }
    \label{tab:ablation_different_design_choices}
\begin{tabular}{llll}
\toprule
Change & NRMSE & SSIM $[\%]$ & PSNR  \\
\midrule
\multirow[t]{3}{*}{Att UNet \&  Mean\eqref{eq:tfm} }  & 0.0261 & 96.04 &  32.30  \\ 
 No Att: Mean & 0.0270 & 95.77 & 31.88 \\
 No Att: Last & 0.0271 & 95.77  &  31.87  \\
No Sparsity Filling & 0.0444 & 90.92 &  27.30 \\
\IB{} + FM* & 0.1029 & 66.83 &  19.97  \\
\midrule
LCI &0.0380&93.50&29.49\\
\bottomrule
\end{tabular}
\end{table}
\\ \noindent \textbf{Insights on design decisions} Table~\ref{tab:ablation_different_design_choices} summarizes the impact of  design choices on ACDC, including an alternative lightweight architecture by replacing the attention mechanism with concatenating time embeddings in the bottleneck, no sparsity filling, and two aggregation methods.
We see that switching to a lighter-weight architecture had only a minor effect on performance.
Future work may explore alternative architectures for the flow network~\eqref{eq:vel_network_def}, either to improve efficiency or further boost performance.
Yet this choice shows the \textit{flexibility of \TFM{}}.
We observe no significant difference between aggregating by the mean or using only the final predicted frame.
Since the model is trained on full flows in both settings, this suggests it learns to predict the target from any context frame.
Crucially, our sparsity filling strategy significantly improves \TFM{}'s performance.
We attribute this due to the fact that filled frames are closer to the target image than zero tensors, resulting in more learnable and stable flow velocities.
This reinforces our core design principle: \textit{\TFM{} focuses on temporal changes, not the entire image distribution}.
\\ \noindent
An important additional finding is that the \IB{} + FM baseline (using \TFM{}'s UNet but a single time channel) performs poorly. 
This occurs even though \TFM{} can handle single context inputs (see Figure ~\ref{fig:front_back_masking}).
Instinctively, both methods should yield similar results at inference, since they receive the same input.
But we believe this discrepancy stems from the training dynamics;
In the \IB{} + FM setting, the model learns the flow over uneven time intervals.
These inconsistent intervals introduce high variance and instabilities in the flows.
In contrast, \TFM{} end-to-end training on the randomly masked sequence yields more information on the temporal spacing, making the predictions more robust against single-input performance.
We suspect that this consistency yields a more stable training regime and enables reliable performance even when reduced to a single input at test time.
\subsection{Future Directions and Limitations}\label{subsec:limitations-and-future-work}
Building on \TFM{}'s core strengths-full resolution flow, end-to-end training, and robust handling of irregular sampling-numerous promising avenues emerge.
None of these expose a fundamental limitation of our method; instead capitalize on its flexibility:
\begin{itemize}
    \item \textbf{Advanced Sparsity Filling} We demonstrated that even a simple nearest-image fill substantially stabilized training (see Table \ref{tab:ablation_different_design_choices}). Future work can explore more sophisticated schemes. This includes learned imputation networks, temporal interpolation priors, which could seamlessly plug into \TFM{}.
    \item \textbf{Explicit Continuous Time Modeling} While our current setting treats $\tau$ as abstract interpolation scalars, this can be naturally extended to model continuous, real-valued time steps. This would allow for more flexible predictions, ideal for clinical workflows. 
    \item \textbf{Stochastic Generation} It is technically simple to include stochastic sampling into \TFM{}. This would allow \TFM{} to sample multiple possible futures. This could be again valuable in the clinical context for risk assessment and planning. On a technical level, this only requires extending the ODE to SDE integration and adding noise during training (see step (6)\ref{alg:tfm}).
\end{itemize}
\paragraph{Limitations}
Several challenges remain, which stem from the realities of longitudinal imaging.
First, truly large-scale, high quality follow-up cohorts remain rare for many diseases. 
Acquiring more multi-timepoint studies can be costly, yet such data are essential for validating disease trajectory models.
Second, our current approach of globally sampling to a set resolution might not be optimal. 
A more localized, patch based strategy could capture finer details better but remains challenging for generative modeling.
Finally, conventional baseline methods struggle when data are scarce, a prominent issue in our setting.
To overcome this limitation, future work might need to leverage large-scale pretrained models, and fine tune them on specific 4D prediction tasks, although such pretrained resources are not yet readily available.
Despite these constraints,  \TFM{} maintains remarkably stable performance.
We hope that this contribution will encourage the acquisition of larger longitudinal datasets and inspire further clinical studies.

\section{Conclusion}
In this paper, we address the challenge of modeling longitudinal medical imaging with sparse and irregular time series by introducing \tfm{} (\TFM{}), a state-of-the-art generative approach for 4D medical image prediction. 
In our datasets, and often in clinical practice, temporal changes constitute only  a small portion of the total image content, relative to inter-patient differences.
\TFM{} leverages this insight by explicitly modeling differences between context and target, an approach which we call \textbf{\dm}.
Through extensive experiments on publicly available datasets, we demonstrate that \TFM{} consistently outperforms prior methods, establishing a new baseline for disease progression modeling. 
While we gratefully acknowledge the public datasets which support this work, advancing spatio-temporal prediction will demand larger cohorts with detailed records of confounding factors, such as surgeries and treatment changes.
\section*{Acknowledgements}
The present contribution is supported by the Helmholtz Association under the joint research school HIDSS4Health – Helmholtz Information and Data Science School for Health.
{\small
\bibliographystyle{ieee_fullname}
\bibliography{references}
}
\clearpage
\appendix
\section{\textit{Method vs Model}}\label{sec:appendix-a_semantics}
In previous sections we mentioned \TFM{} as a baseline \textit{method}.
To further disambiguate semantic definitions, we define the following;
\textbf{Architecture} is defined here as the actual neural network. 
That is the functional output between input of the network and the output. 
This seems redundant, but important in comparison.
We define \textbf{Model}  as the input and especially output space. The best example here is diffusion vs.\, flow matching;
Both can be done via the same network $f_\theta$, but in the case of diffusion, the input is a noisy sample, and the output is the noise.
Whereas for flow matching, the network receives $\{X_\tau,\tau\}$, and  predicts $u_\tau$. 
We say the diffusion network \textit{models} the noise, and the flow matching network \textit{models} the velocity.
\\
\noindent

\section{Datasets}

\begin{figure}[htbp]
    \centering
    \includegraphics[width=0.45\textwidth]{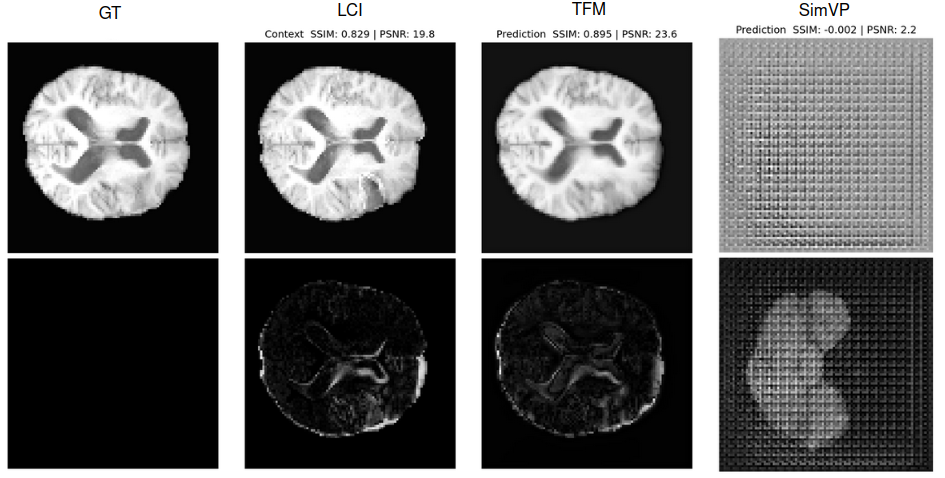}
    \caption{\textbf{Example prediction from the Lumiere Dataset}: 
    Comprarison of prediction results for \IB{}, \TFM{} and SimVP.
    Ground truth (GT) is shown in the first column. The second row visualizes the per-method absolute prediction error.
    While \TFM{} reconstructs with decent fidelity compared to \IB{},SimVP fails to generalize to this case. 
    }
\end{figure}
\begin{figure}[htbp]
    \centering
    \includegraphics[width=0.45\textwidth]{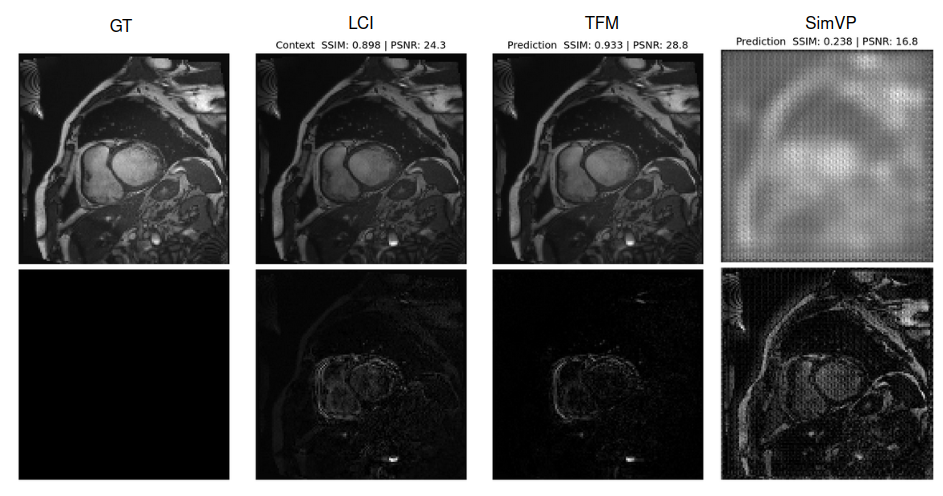}
    \caption{\textbf{Example prediction from the ACDC dataset}: 
    Comparison of \IB{}, \TFM{} and Simvp against the GT on the ACDC dataset. Despite having a high \IB{} performance, \TFM{} improves on it. 
    Note that the absolute differences are compact, similar to \IB{}, but less pronounced.
    SimVP fails to recover the finer spatial details, but it can reconstruct the broad details.
    The second row shows the absolutes from the residuals, highlighting each method's reconstruction error.
    }
\end{figure}
\begin{figure}[htbp]
    \centering
    \includegraphics[width=0.95\linewidth]{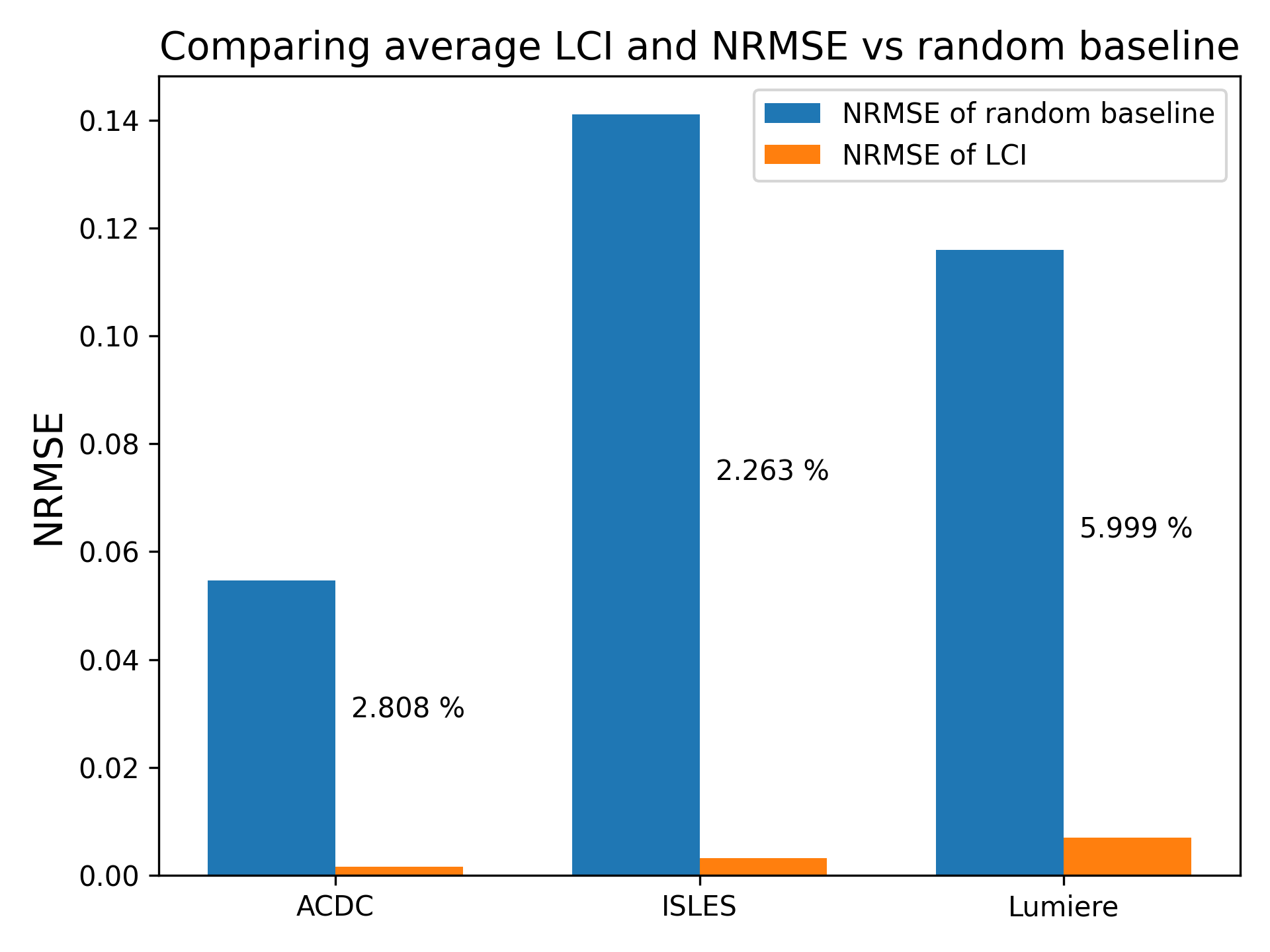}
    \caption{
    \textbf{\IB{} is \textit{close} to the Target}: We report two $NRMSE$ values: one between \IB{} and $I_\text{target}$, and another between target and a random image.
    Percentages indicate the ratio of the \IB{} error to the random image baseline error. For reference, the $NRMSE$ between two random patients in the ACDC dataset is approximately $0.0287$, which is about half the error of the random baseline.
    }
    \label{fig:ratios_mse_lci}
\end{figure}

\begin{figure*}[ht]
\centering
\includegraphics[width=3.0cm,height=3.0cm]{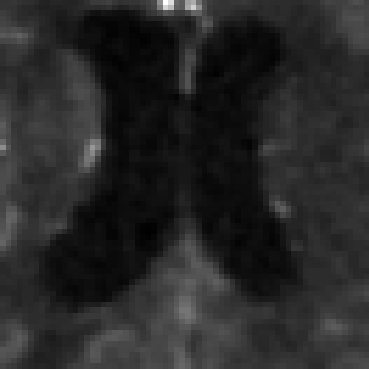}\hspace{0.2cm}%
\includegraphics[width=3.0cm,height=3.0cm]{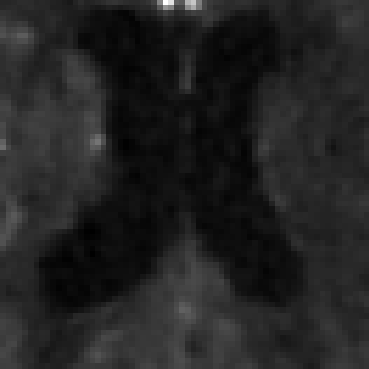}\hspace{0.2cm}%
\includegraphics[width=3.0cm,height=3.0cm]{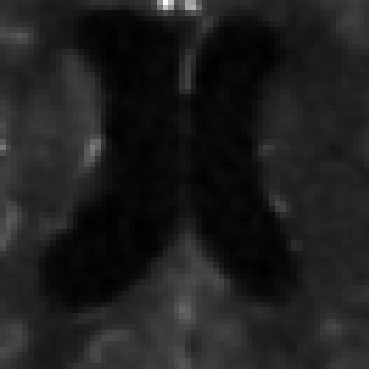}\hspace{0.2cm}%
\includegraphics[width=3.0cm,height=3.0cm]{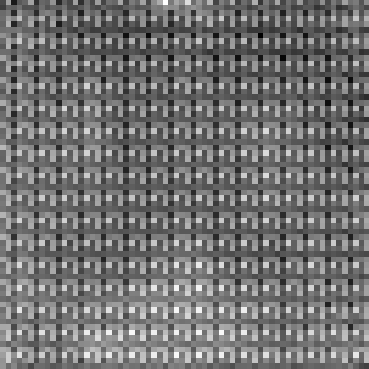}\\[0.3em]
\includegraphics[width=3.0cm,height=3.0cm]{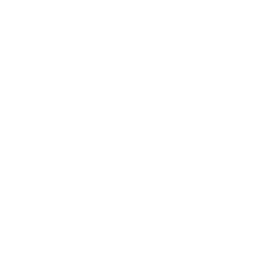}\hspace{0.2cm}%
\includegraphics[width=3.0cm,height=3.0cm]{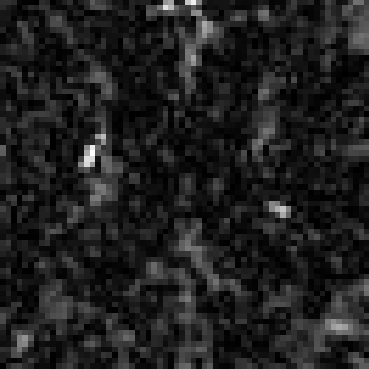}\hspace{0.2cm}%
\includegraphics[width=3.0cm,height=3.0cm]{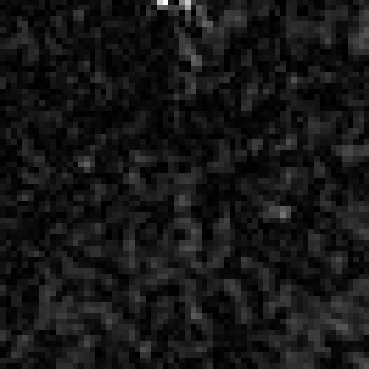}\hspace{0.2cm}%
\includegraphics[width=3.0cm,height=3.0cm]{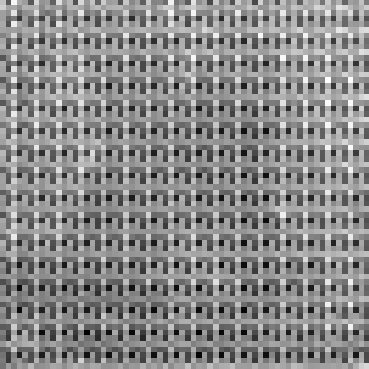}\\[0.3em]
\caption{\textbf{Zoomed-in prediction Examples from the  ISLEs Dataset}: 
Visual comparison of predictions from \IB{}, \TFM{}, and SimVP, alongside the Ground Truth, \textit{focusing on a high-resolution crop for clarity}. 
The second row shows the absolute values from the residuals.
\TFM{} preserves the spatial quality, while SimVP struggles to recover the underlying details. 
This shows that even in fine details. \TFM{} maintains spatial structures.
Furthermore, in the residuals, we can see that they are lower and less noisy.
For this specific patch, the $NRMSE$ are in order: $0.0028, 0.0016, 0.0751$.
}
\label{fig:seven-grid}
\end{figure*}

\begin{table}[ht]
\centering
\begin{tabular}{cc}
    \toprule
\textbf{NFEs} & \textbf{SSIM}  \\
\hline
1 & 0.956645\\
5 & 0.959890 \\
10 & 0.959926 \\
25 & \textbf{0.959954} \\
50 & 0.959951  \\
100 & 0.959926 \\
150 & 0.959879  \\
200 & 0.959877  \\
300 & 0.959884 \\
400 & 0.959920  \\
    \bottomrule
\end{tabular}
\caption{
\textbf{Evaluating $SSIM$ vs. Number of Function Evaluations:}
We evaluate how the number of function evaluations (NFEs) affects $SSIM$ performance on one ACDC validation set. 
$SSIM$ increases with more evaluations and peaks at $25$ NFEs, after which it plateaus. 
However, the improvement becomes marginal after beyond just $5$ NFEs.
}
\label{tab:ssim_vs_memory}
\end{table}
In table \ref{tab:ssim_vs_memory} we ablate the amount of number of function evaluations. We note that for a single NFE the performance significantly drops. For a trade-off we chose 10 as the integration steps for all datasets.

\section{Toy Example - Difference Modeling}\label{sec:appendix_paradox}
To clarify why modeling differences can be advantageous in low-change environments (even when full image reconstruction is limited), we present a simplified toy example illustrating a resolution-performance paradox.
To show how limited resolution and bounded changes can impact error metrics, consider the following setting:
Assume an $8 \times 8$ checkerboard pattern where each pixel alternates between $0$ and $1$. 
In the center, a $4 \times 4$ patch undergoes a change, specifically within a $2 \times 2$ region. 
For simplicity, let $I_0$ contain two black squares in the center, and $I_1$ contain three. 
Suppose a perfect longitudinal model captures the central change but operates at a coarse resolution of $2 \times 2$.
Since it cannot represent the high-frequency checkerboard pattern, its best prediction is a uniform value of $0.5$ across the image. This yields a total MSE of $12/64$, but an \IB{} MSE of only $4/64$.
Now consider the difference image $I_1 - I_0$, which contains a single $2 \times 2$ black square and zeros elsewhere. 
 The same low-resolution model can now perfectly represent this difference, achieving an MSE of $0$, despite lacking the resolution to represent the full images individually.
 This illustrates how \dm can resolve the apparent paradox where a model with limited spatial capacity still performs well under certain metrics. 
 While this does not fully explain the behavior of \TFM{}, it provides intuition for how modeling the difference yields strong starting conditions, and how methods can benefit from this formulation.

\section{Further Experimental Settings}
Models were trained for $500$ epochs.
All methods are implemented using the AdamW optimizer, with cosine annealing learning rate, and batch size 4.
We utilized cosine annealing, as well as a warm-up scheduler for $10\%$ of the total epochs, as well as a gradient clipping of magnitude $1$. 
\\
\textbf{\TFM{} network details}
For the experiments we used a feature size of $32$, and a channel multiplication per layer of $(1,1,2,4)$, with one res block per layer. 
The attention resolution was set to $16$.
For anything else, the default parameters of the UNet from \cite{tongTorchCFM2025} was used.

\end{document}